\documentclass{article}
\usepackage{spconf,amsmath,graphicx}
\usepackage{algorithm}
\usepackage{algorithmic}
\usepackage{cite}
\usepackage{algorithmic}
\usepackage{graphicx}
\usepackage{subfig}
\usepackage{textcomp}
\usepackage{xcolor}
\usepackage{multirow}
\usepackage{url}

\title{LIDAR ICPS-NET: INDOOR CAMERA POSITIONING BASED-ON Generative Adversarial Network for RGB TO POINT-CLOUD translation}
\name{Ali Ghofrani$^1$, Rahil Mahdian Toroghi$^1$, Seyed Mojtaba Tabatabaie$^2$, Seyed Maziar Tabasi$^2$}
\address{
	$^1$ Faculty of Technology and Media Engineering, Iran Broadcasting University (IRIBU), Tehran, Iran\\
	$^2$ CEO/CTO at Alpha Reality, AR/VR Solution Company \\
	alighofrani@iribu.ac.ir, mahdian.t.r@gmail.com, \{smtabatabaie,m.tabasi\}@alphareality.io}
\begin{document}
%
\maketitle
\begin{abstract}
Indoor positioning aims at navigation inside areas with no GPS-data availability, and could be employed in many applications such as augmented reality, autonomous driving specially inside closed areas and tunnels. In this paper, a deep neural network based architecture has been proposed to address this problem. In this regard, a tandem set of convolutional neural networks, as well as a Pix2Pix GAN network have been leveraged to perform as the scene classifier, scene RGB image to point cloud converter, and position regressor, respectively. The proposed architecture outperforms the previous works, including our recent work, in the sense that it makes data generation task easier and more robust against scene small variations, whilst the accuracy of the positioning is remarkably well, for both Cartesian position and quaternion information of the camera. 

\end{abstract}
\vspace{-4mm}
\section{Introduction}
\label{sec:intro}
Global positioning system is a problem, which has been contributed using navigation systems, and GPS satellites. The indoor positioning, on the other hand is still challenging task due to the fact that inside covered areas with no GPS signal available, image processing tasks are the only solutions to be resorted (e.g., SIFT and SURF). These methods are not very accurate~\cite{sattler2016efficient}. The main reason is the existence of several identical patterns inside the buildings, which could easily fool the positioning system.

The first data-driven approach using convolutional neural networks (CNN) was POSENET ~\cite{kendall2015posenet}, which could work for a limited open area. Further, a geometry-aware system was proposed for camera localization which incorporated perceptual and temporal features to improve the precision, ~\cite{henriques2018mapnet}. However,  both these methods were applicable in outdoor positioning. In most traditional indoor positioning systems, which do not involve wireless means \cite{yang2015wifi}, the depth-assisted camera is necessary to be used \cite{zhang2018real}, which is not always available in real-world scenarios, such as mobile handsets.

 The first indoor positioning system using deep neural networks, was proposed by the authors of this paper \cite{ghofrani2019icps}, through scanning of the desired area segments using photogrammetry method. A classifier is then trained by a CNN structure (i.e. EfficientNet \cite{tan2019efficientnet}), and followed by a MobileNet CNN structure \cite{sandler2018mobilenetv2}, which has already been trained to perform as a regressor. This structure could achieve a remarkable precision result for the Cartesian position and quaternion information of the camera \cite{ghofrani2019icps}. The remaining challenge of the previous work is that, generation of such a huge amount of RGB data for training the deep neural network is an overwhelming task. Moreover, for the case in which the area is subject to small changes, then the RGB based data is no longer trustable and the output of the previous system is not robust, at all.
 
 A solution to the aforementioned problem would be to generate a point cloud data using a LiDAR system rather than RGB cameras, which is both easier and more robust.
 
In this work, we extended our research to investigate whether it would be possible for our regressor-CNN to be driven by a point-cloud data, rather than the RGB image.
Wang et al. in \cite{wang2019pseudo}, showed that it would be possible to detect the object using its associated point-cloud data. On the other hand, Shi et al. \cite{Shi_2019_CVPR}, showed that it is possible to render the point-cloud data into associated images using GAN neural networks \cite{pix2pix2016}.

\begin{figure}[!h]
	\centering
	\includegraphics[width=1\linewidth, height=.43\textheight]{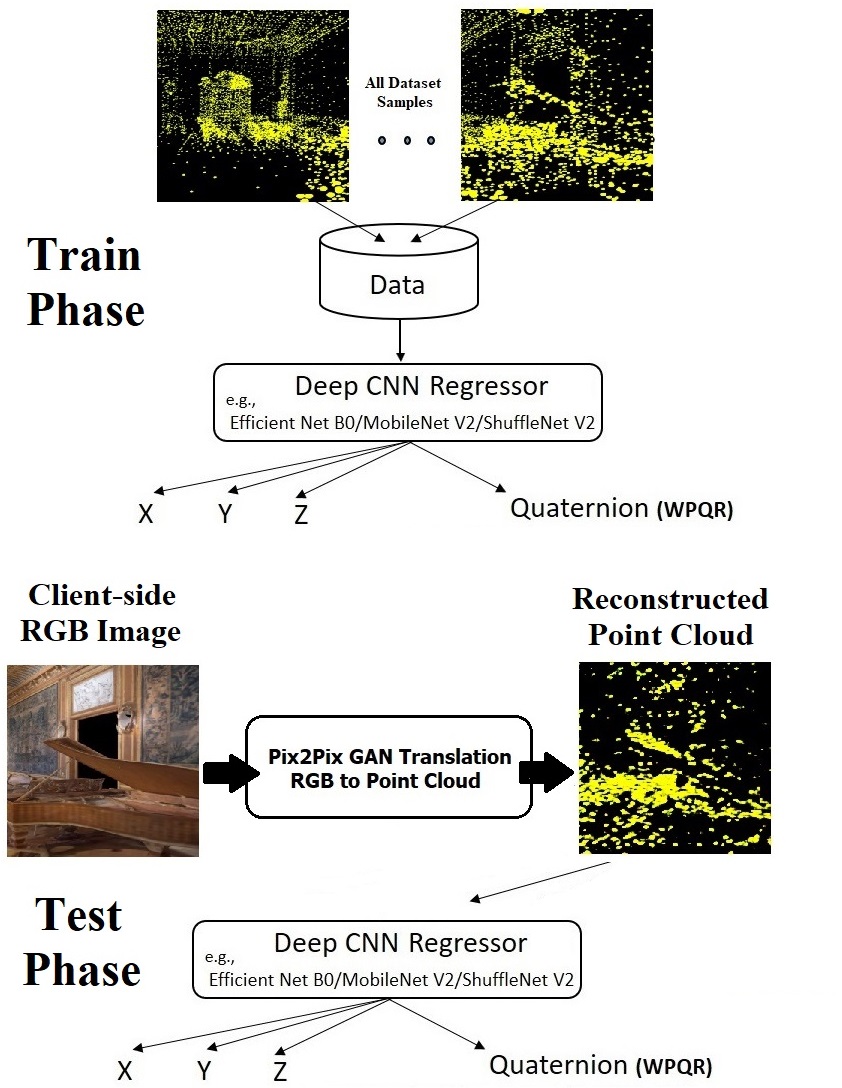}
	\caption{A big-picture of the LiDAR-based indoor positioning. }
	\label{fig:phases}
	\vspace{-5mm}
\end{figure}

 Following these two works, as illustrated in figure~\ref{fig:phases}, the CNN-regressor is trained by the point-cloud data instead of the RGB image. Moreover, due to the fact that the clients normally have access to only RGB images on their mobile handsets, therefore we need a transformer which converts the RGB data into its associated point-cloud data which we perform it using a Pix2Pix GAN neural network to achieve this mapping. This enables the training procedure to be performed much easier than our previous work, and further within small environmental changes the model could perform more robust than before. These are explicitly the novelties of our work. 
 
\section{The Proposed Framework}
\begin{figure}[!b]
 	\centering
 	\includegraphics[width=0.8\linewidth, height=.35\textheight]{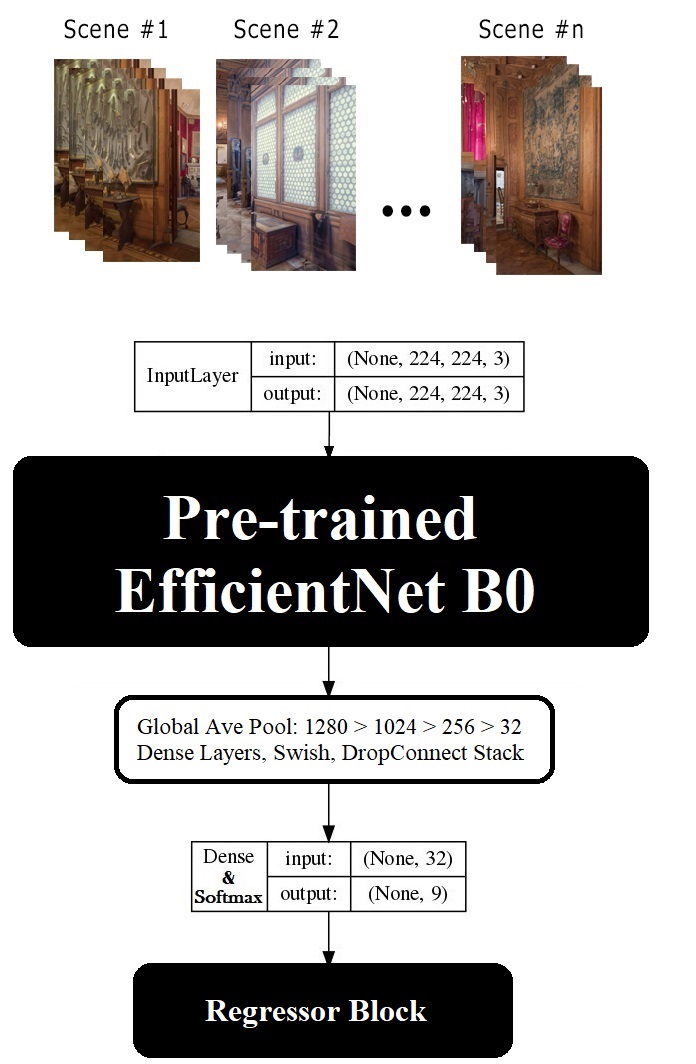}
 	\caption{Scene classifier based on EfficientNet B0.}
 	\label{fig:class}
 	\vspace{-3mm}
 \end{figure}

\vspace{-2mm}
\label{sec:format}
Regarding our previous work \cite{ghofrani2019icps}, the following steps should be taken in a sequence:
 1) The input images of the clients should be given to a classifier in order to determine the associated scene. Segmentation of the desired environment into scenes could be optional. However, when we decide about the number of scenes we have to fix it, and the classifier should be trained based on that. The structure of this scene classifier, which is an EfficientNet B0 \cite{tan2019efficientnet}, is depicted in figure~\ref{fig:class}. 
 \begin{figure*}[!t]
 	\centering
 	\includegraphics[width=1.02\linewidth, height=.23\textheight]{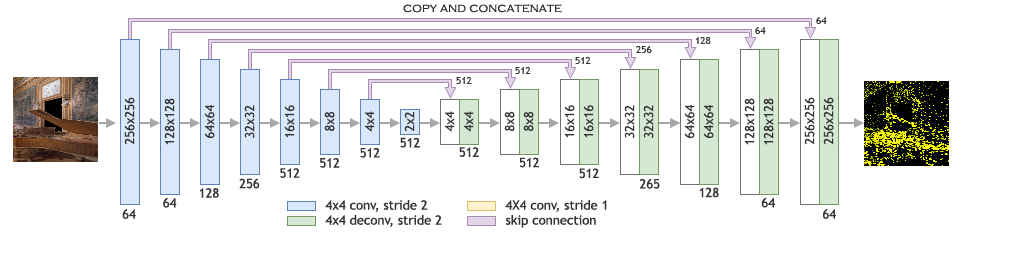}
 	\caption{Image-to-image (RGB-2-Pointcloud) translation, using Pix2Pix GAN ~\cite{pix2pix2016}}
 	\label{fig:unet}
 \end{figure*}
  \begin{figure} [!b]
  	\centering
  	\includegraphics[width=0.98\linewidth, height=.16\textheight]{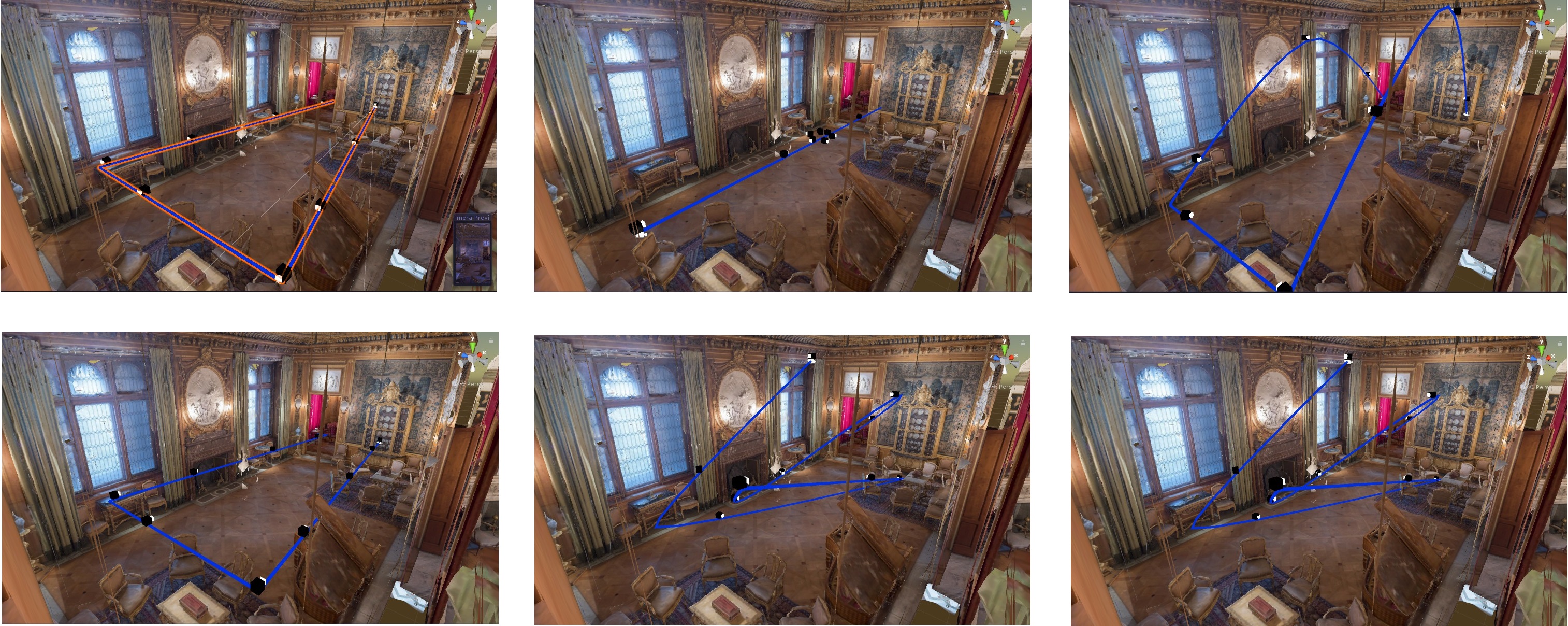}
  	\caption{Sequences of Camera movements for each scene}
  	\label{fig:dataset}
  	\vspace{-3mm}
  \end{figure}
  2) When the classifier determines the scene, the RGB image should be converted to its associated point-cloud using a Pix2Pix UNET-based GAN network \cite{ronneberger2015u}. 3) This generated point-cloud data, would be fed into the CNN-based regressor which has been trained based upon its associated scene.
 \begin{figure}[!h]
 	\centering
 	\includegraphics[width=0.85\linewidth, height=.55\textheight]{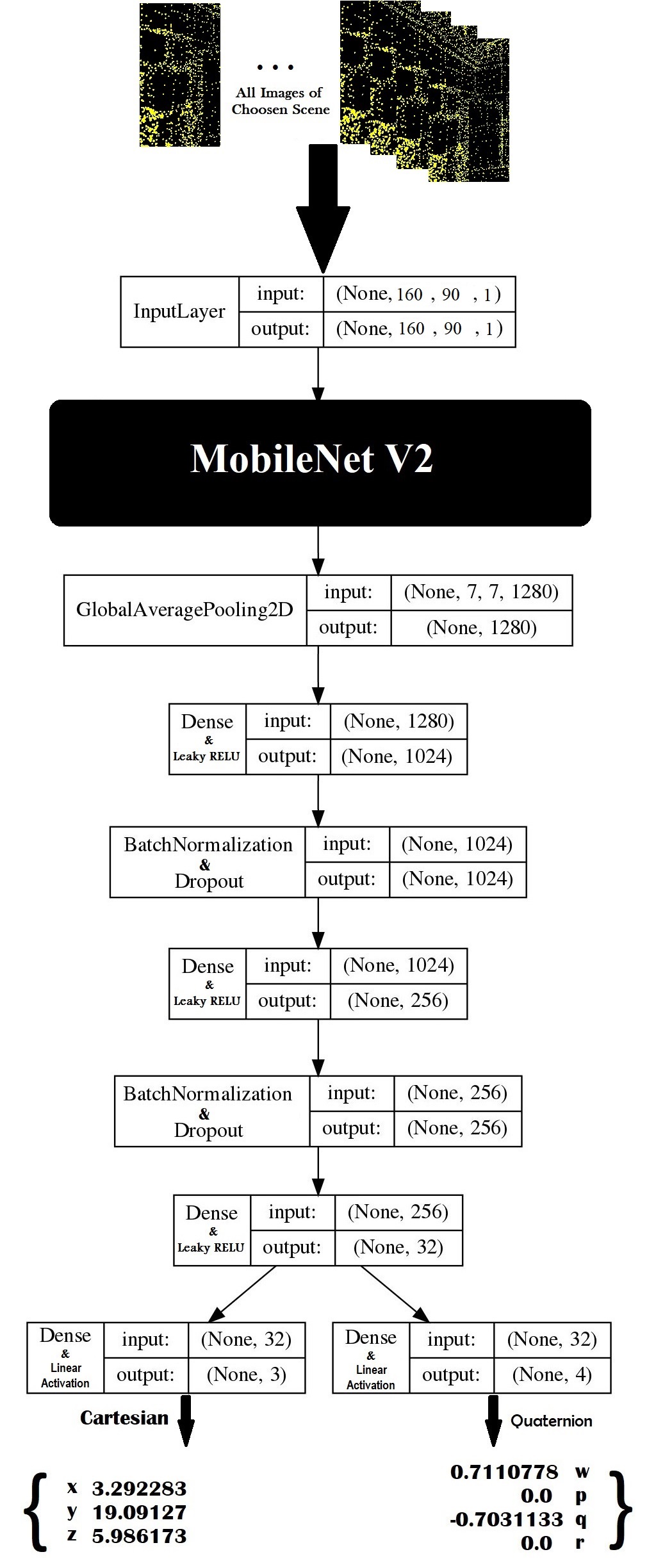}
 	\caption{MobileNet V2, as the regressor trained by the point-cloud dataset}
 	\label{fig:final-regressorasas}
 	\vspace{-4mm}
 \end{figure} 
 Based on the above procedure, we need to primarily train a UNET-based GAN network \cite{ronneberger2015u}, to perform the mapping of RGB images into point-cloud data. For this purpose, using a small amount of data samples which contain the pairs of RGB images, and their associated and compatible point-cloud data we could train the network. This network is depicted in figure~\ref{fig:unet}.
 
 Next, we need to train the regressor network, which is supposed to get the generated point-cloud data as the input and estimate the 7 values of Cartesian and Quaternion information as the output. This CNN-regressor (based on MobileNet V2) is depicted in figure~\ref{fig:final-regressorasas}.
 \section{Experiments and Analytics}
 The hardware being used for the present work, is GTX 1080-NVIDIA, on a core i7 Cpu Intel 7700, with 32 GB RAM. Tensorflow 1.13.1 has been used with CUDA 10.1, and Keras 2.2.4 softwares are the platforms to implement the tasks.
 
 Since there were no available data containing the RGB and associated point-cloud, we generated this dataset from the freely available 3D scanned images of the Hallwyl museum in Stockholm ~\cite{ghofrani2019hallwylmuseum}. We sampled from this 3D model using the Unity software, and the normalized outputs are saved in our generated dataset\footnote{\url{https://mega.nz/#F!FE9HFCLS!vHH7vqEd5PAFF-ItGR44ww}}$^,$\footnote{\url{https://drive.google.com/drive/folders/1Q2QaiQejigriIaFxn7G9csEXD6OEkYvm}}. More than $500,000$ pure data samples are generated from all the scenes using different regimes for the camera, depicted in figure~\ref{fig:dataset}. The equivalent point-cloud data for each of the image samples are created. 
 
\begin{figure}[!t]
 	\centering
 	\includegraphics[width=0.98\linewidth, height=.27\textheight]{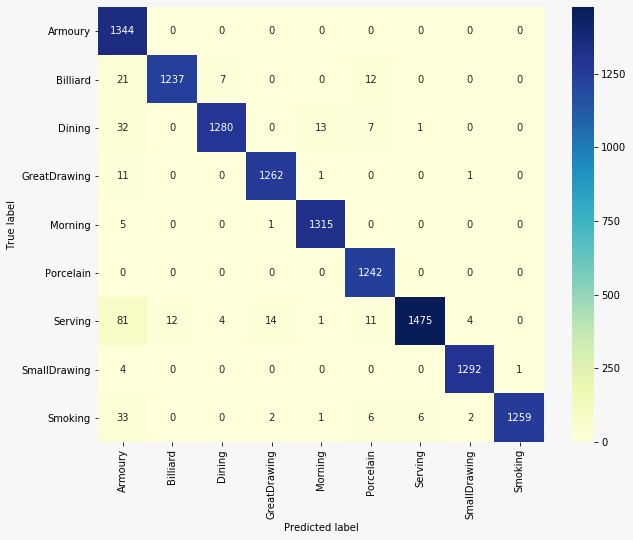}
 	\caption{The confusion matrix for the classification of the scenes through EfficientNet.}
 	\label{fig:test_sample_ConfusionMatrix}
 	\vspace{-3mm}
 \end{figure}

 In order to create the point-clouds, inside the Unity software we have modified the mesh descriptor of the environment mesh from the surface shader to geometry shader, in which the mesh vertexes are demonstrated using the points. Thus, for each RGB image the equivalent point-cloud data has been created. 
 Since the GAN network training, requires some RGB and associated point-cloud data pairs, and the scene classifier also needs to be trained on the scenes through RGB images this may give the wrong impression that the RGB images are again under usage. However, the amount of RGB images which could be employed for the GAN network is sufficient to train the classifier network, as well. This has been investigated and the result confusion matrix has been depicted in figure~\ref{fig:test_sample_ConfusionMatrix}. 
 
 For the classifier, the loss function being used is the categorical cross-entropy, and the model is monitored toward maximizing the validation accuracy. 
 
 \begin{figure}[!h]
 	\centering
 	\includegraphics[width=0.9\linewidth, height=.4\textheight]{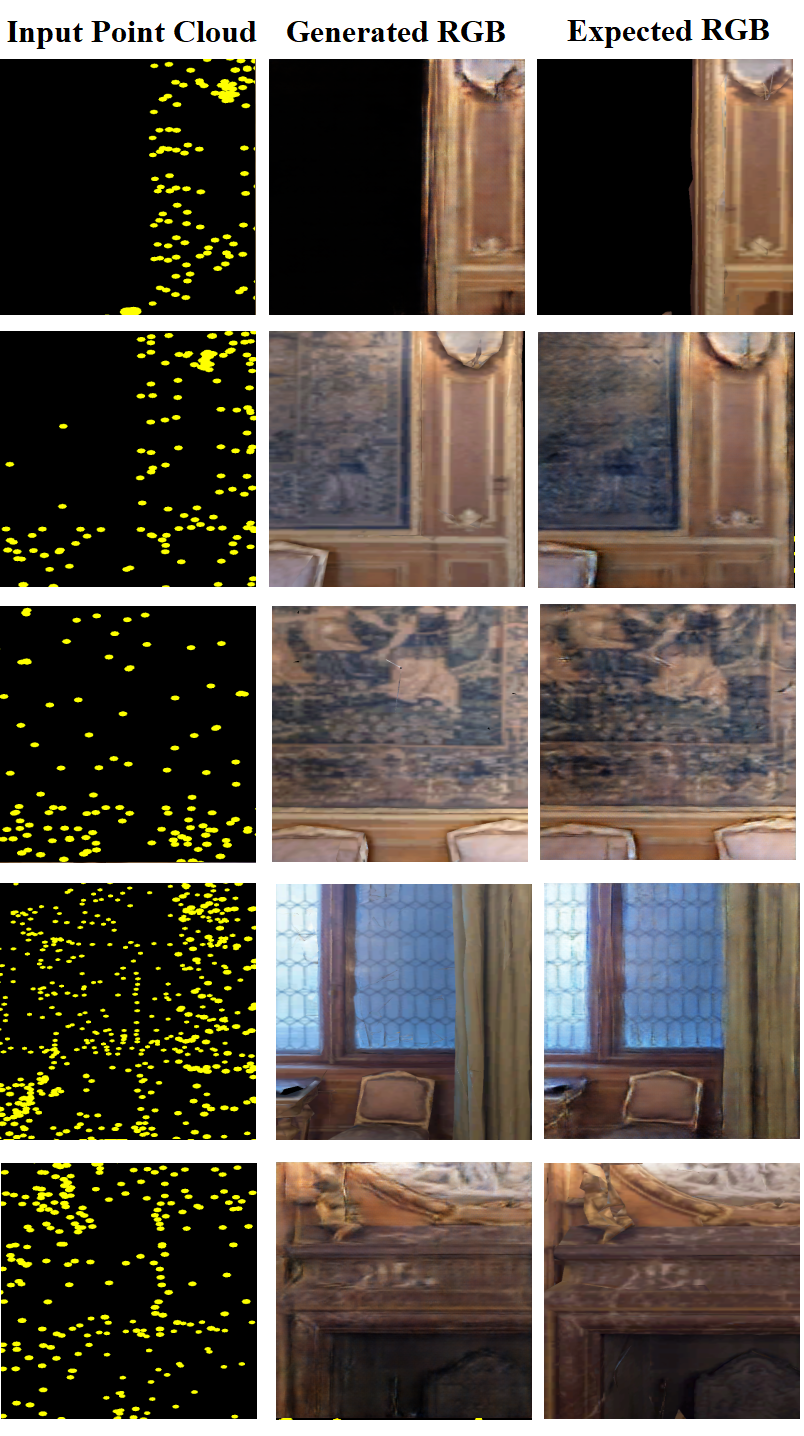}
 	\caption{(Left to right) input Point cloud, generated RGB-GAN output, and the ground truth RGB.}
 	\label{fig:pc2rgb}
 	\vspace{-7mm}
 \end{figure}
 \begin{figure}[!b]
 	\centering
 	\includegraphics[width=1\linewidth, height=.135\textheight]{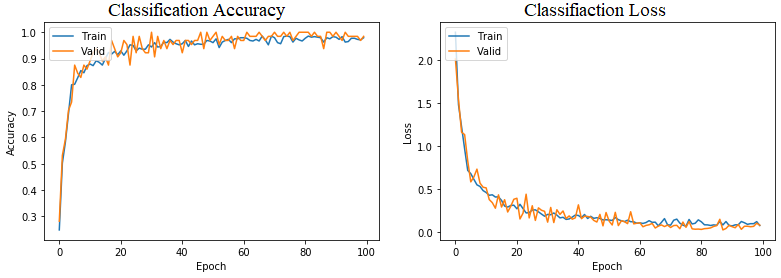}
 	\caption{Classification accuracy (left), and loss (right) based on the categorical cross-entropy.}
 	\label{fig:new_loss_status}
 \end{figure}
 In order to achieve the optimum performance, the drop-connect is employed to avoid overfitting \cite{wan2013regularization}. In addition, the swish as a SOTA activation function has been used, as the state-of-the-art \cite{ramachandran2017swish}. 
 
\begin{figure}[!t]
\centering
\includegraphics[width=1\linewidth, height=.13\textheight]{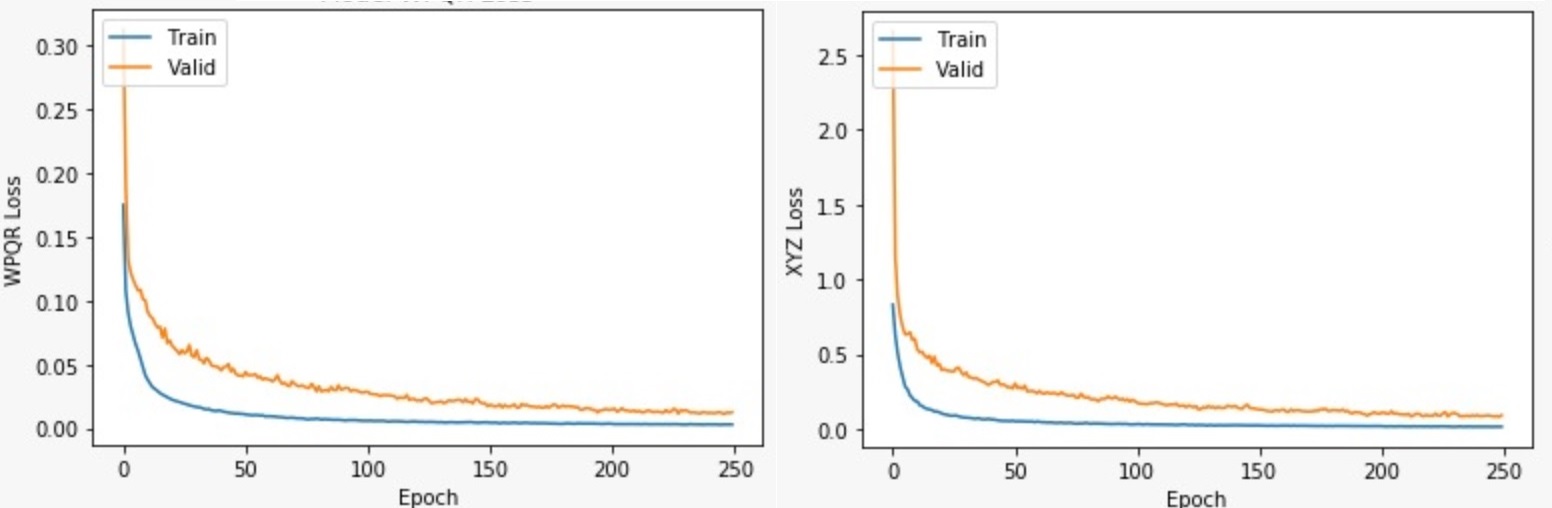}
\caption{(left) Quaternion loss, (right) Cartesian Loss. Losses are to be scaled using the scale factor in the loss function.}
\label{fig:regressor_losses}
\end{figure}

 To train the regressors, since the input dataset is point-cloud, it is not possible to use the imageNet-based training parameters, in a transfer learning procedure. Therefore, the entire training of the regressors has been performed from scratch via Xavier weight initializing technique \cite{DBLP:journals/corr/Kumar17}. The loss changing diagram has been depicted in figure~\ref{fig:regressor_losses}.

The loss function should be chosen as in \cite{ghofrani2019icps}. This loss function is, as follows
\begin{equation}
 loss = ||P-\hat{P}||_2 + \frac{1}{\beta} ||\hat{Q}-\frac{Q}{||Q||} ||_2
\end{equation}
where $P = [x, y, z]$ is the position data vector, Q is the quaternion information, and $\beta$ is the scale factor to make a balance between estimating the position and the quaternion. 
\begin{figure}[!t]
	\centering
	\includegraphics[width=0.9\linewidth, height=.41\textheight]{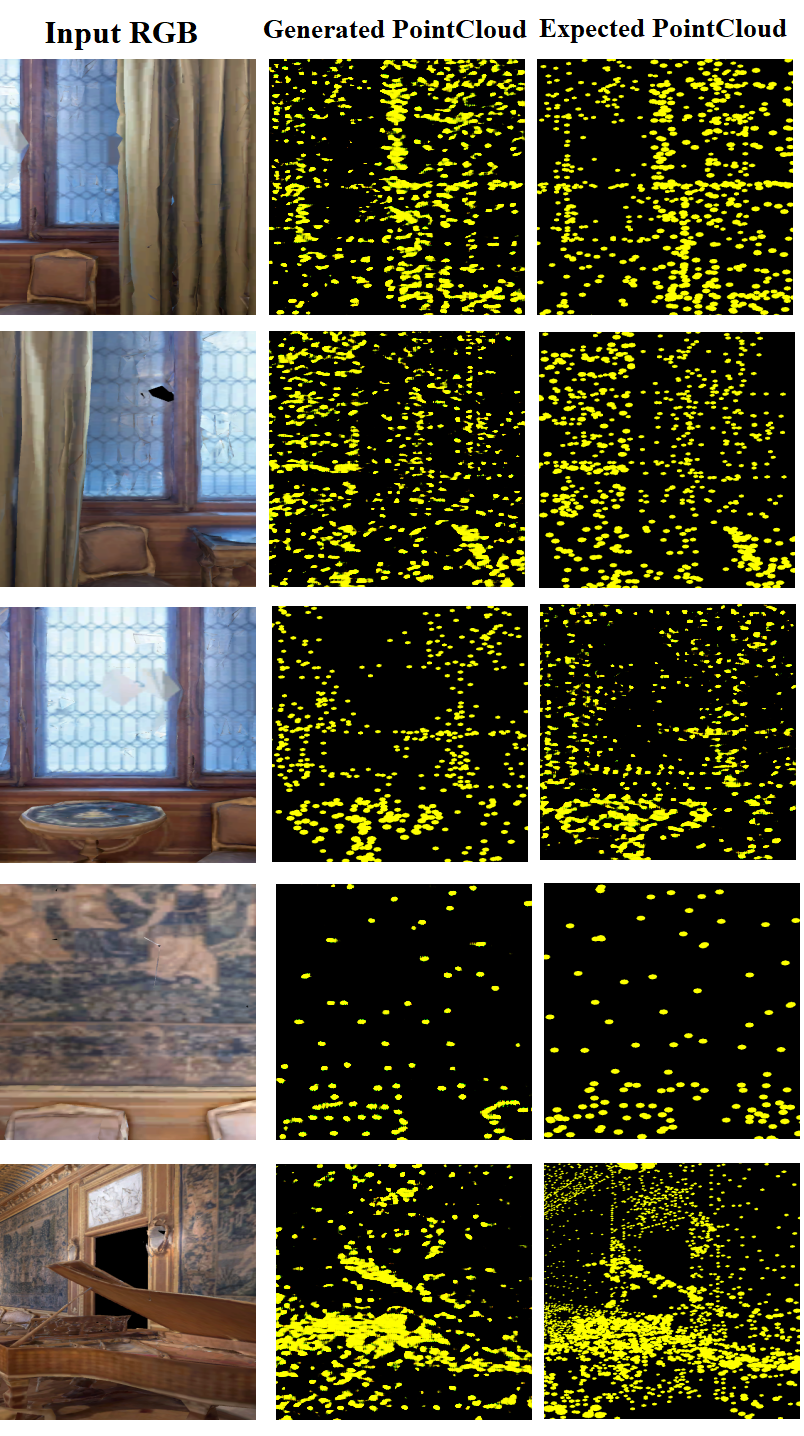}
	\caption{(Left to right) RGB input data, generated point cloud-GAN output, and the ground truth point cloud.}
	\label{fig:rgb2pc}
	\vspace{-3mm}
\end{figure}
The GAN training is based on the RGB-2-Point cloud data, which has been generated, as mentioned before. A sample of this data has been depicted in figure~\ref{fig:rgb2pc}.
In a further investigation, we turned the GAN to work as a point cloud to RGB converter. Interestingly, the same network could perform quite well, as depicted in figure~\ref{fig:pc2rgb}.


\begin{table} [!b]
	\centering
	\caption{The regression error, for the position vector (X;Y;Z), and the camera Quaternion, over the test set (Unseen data) }
	\vspace{-2mm}
	\begin{tabular}{l|l|l|l|l|} 
		\cline{2-5}
		&\footnotesize \textbf{X}-position &\footnotesize \textbf{Y}-position &\footnotesize \textbf{Z}-position & \textbf{\footnotesize Quaternion }  \\ 
		\hline
		\multicolumn{1}{|l|}{\footnotesize\textbf{Error Value}} &\footnotesize \textbf{0.019 m}   &\footnotesize \textbf{0.027 m}   &\footnotesize \textbf{0.0073 m}   &\footnotesize \textbf{~ ~0.0096}    \\
		\hline
	\end{tabular}
	\vspace{-8mm}
	\label{tbl:error_final}
\end{table}

\section{Conclusion}
An indoor position system has been proposed in this paper, based on a supervised deep network structure. The goal of the system is to achieve a high accuracy of the Cartesian (X,Y,Z) position and the camera quaternion, while being robust against environmental changes and object movements. A CNN-based classifier is used to identify the scene from the environment based on the client's input RGB image. A GAN network has already been prepared to convert the RGB images into point cloud data which is easier available and more robust against variations of the scene background. 
The regressor CNNs are trained only based on the point clouds. The results of the experiments showed a remarkable achievement in positioning whilst making the entire procedure of our previous work much easier to be performed.

\vspace{-2mm}

\bibliographystyle{IEEEbib}
\bibliography{Ghofrani}

\end{document}